\documentclass[letterpaper, 10 pt, conference]{ieeeconf}  

\IEEEoverridecommandlockouts                              

\usepackage{amsmath}
\usepackage{amsfonts}
\usepackage{graphics} 
\usepackage{graphicx}
\usepackage{todonotes}
\usepackage{array}
\usepackage{multirow}
\usepackage{verbatim} 
\usepackage{mathtools}
\usepackage{amssymb}
\usepackage{url}
\usepackage{tikz}
\usepackage{hyperref}
\usepackage{subfig}
\usepackage{float}
\usepackage{amsmath}
\usepackage{amsfonts}

\usepackage{algorithm}
\usepackage{algpseudocode}
\usepackage{eso-pic}
\newcommand\AtPageUpperMyright[1]{\AtPageUpperLeft{
 \put(\LenToUnit{0.5\paperwidth},\LenToUnit{-1cm}){
     \parbox{0.5\textwidth}{\raggedleft\fontsize{9}{11}\selectfont #1}}
 }}
\newcommand{\conf}[1]{
\AddToShipoutPictureBG*{
\AtPageUpperMyright{#1}
}
}

\conf{This Article has been accepted in IEEE 18th International Conference on Control, Automation, Robotics and Vision (ICARCV), ICARCV 2024,  12 – 15 December 2024, Dubai, UAE. Kindly cite accordingly}

\title{\LARGE \bf
Active Collaborative Visual SLAM exploiting ORB Features
}

\author{Muhammad Farhan Ahmed$^{1}$, Vincent Frémont$^{1}$ and Isabelle Fantoni$^{2}$
\thanks{}
\thanks{$^{1}$Muhammad Farhan Ahmed and Vinent Frémont are with LS2N, Ecole Centrale de Nantes (ECN), 1 Rue de la Noë, 44300 Nantes, France
        {\tt\small Muhammad.Ahmed@ec-nantes.fr}, 
        {\tt\small vincent.fremont@ec-nantes.fr}}%
\thanks{$^{2}$Isabelle Fantoni is with LS2N, Centre National de la Recherche Scientifique (CNRS), 2 Chemin de la Houssinière, 44322 Nantes, France
        {\tt\small isabelle.fantoni@ls2n.fr}}%
}

\begin{document}
\maketitle
\thispagestyle{empty}
\pagestyle{empty}

\begin{abstract}
In autonomous robotics, a significant challenge involves devising robust solutions for Active Collaborative SLAM (AC-SLAM). This process requires multiple robots to cooperatively explore and map an unknown environment by intelligently coordinating their movements and sensor data acquisition. In this article, we present an efficient visual AC-SLAM method using aerial and ground robots for environment exploration and mapping. We propose an efficient frontiers filtering method that takes into account the common IoU map frontiers and reduces the frontiers for each robot. Additionally, we also present an approach to guide robots to previously visited goal positions to promote loop closure to reduce SLAM uncertainty. The proposed method is implemented in ROS and evaluated through simulations on publicly available datasets and similar methods, achieving an accumulative average of 59\% of increase in area coverage.
\end{abstract}

\section{INTRODUCTION}
\label{sc: introduction}

Current research is increasingly focusing on Active Collaborative SLAM (AC-SLAM), which leverages multiple robots working in unison. This collaborative method offers significant advantages, such as faster terrain mapping and robust operation in dynamic scenarios. However, the deployment of multiple robots introduces challenges like coordination, resource management, and sensor data integration. Moreover, merging individual robot contributions into a unified map poses complex computational and algorithmic challenges.
Here, we propose an implementation of a visual AC-SLAM algorithm using heterogeneous agents and extend the work in \cite{julio1} to a multi-agent system, where a team of ground robots and aerial robots (UAVs) collaboratively map and explore the environment. To achieve this aim, we propose a method that exploits ORB \cite{orb} feature information from different viewpoints of UAVs and ground robots into a common merged map, we detect frontiers in local and overlapped ground robot maps and propose a frontier filtering method that encourages the spread of agents and exploration of the environment. Additionally, we also present a re-localization scheme to guide robots to previously visited goal positions to favor loop closure.   

The subsequent sections are organized as follows: Section \ref{sc: related_work} provides a review of related work, Section \ref{sc: methodology} explains the methodology of the proposed approach, Section \ref{RESULTS} shows the simulation results, finally we conclude Section \ref{sc: conclusions} summarizing our contributions and prospects for future work. Throughout this article, we will use the words \textit{robots} or \textit{agents} interchangeably, and the same applies to \textit{frontiers} and \textit{points}, as they imply the same meaning in the context. 

\section{RELATED WORK}
\label{sc: related_work}

\subsection{Active Collaborative SLAM}

In AC-SLAM, the frontier detection and uncertainty quantification approaches debated in A-SLAM \cite{farhan1} are also applicable with additional constraints of managing computational and communication resources, and the ability to recover from network failure. The exchanged  parameters are entropy \cite{dariothesis}, localization info \cite{AC15}, visual features \cite{AC16}, and frontier points. The authors of \cite{AC2}, incorporate these multirobot constraints by adding the future robot paths while minimizing the optimal control function which takes into account the future steps and observations and minimizing the robot state and map uncertainty and adding them into the belief space (assumed to be Gaussian).

 \cite{AC17} presents a decentralized method for a long-planning horizon of actions for exploration and maintains estimation uncertainties at a certain threshold. The active path planner uses a modified version of RRT* and an action is chosen that best minimizes the entropy change per distance traveled. The main advantage of this approach is that it maintains good pose estimation and encourages loop-closure trajectories. An interesting solution is given by a similar approach to the method proposed by \cite{AC18} using a Relative Entropy (RE) optimization method which integrates motion planning with robot localization and selects trajectories that minimize the localization error and associated uncertainty bound. A planning-cost function is computed, which includes the uncertainty in the state in addition to the state and control cost.

\subsection{Frontiers-based exploration}
 Frontiers play a pivotal role in augmenting the precision of robot localization by enabling intelligent exploration and data acquisition strategies, effectively reducing uncertainty, and enhancing the map building and localization processes. In \cite{7125079} each frontier is segmented, a trajectory is planned for each segment, and the trajectory with the highest map segment covariance is selected from the global cost map. The work presented in \cite{dariothesis} uses frontier exploration for autonomous exploration a utility function based on Shannon's and Renyi entropy is used for the computation of the utility of paths. The method described by \cite{CS14} uses a cost function that is somewhat similar to \cite{AC15}, it takes into consideration the discovery of the target area of a robot by another member of the swarm and switches from a frontier to a distance based navigation function to guide the robot toward the goal frontier.
  Frontiers-based coverage approaches in \cite{AC8} divide the perception task into a broad exploration layer and a detailed mapping layer, making use of heterogeneous robots to carry out the two tasks while solving a Fixed Start Open Traveling Salesman Problem (FSOTSP). Once a frontier has been identified, the robot can use path planning algorithms to reach it and maximize the exploration while minimizing its SLAM uncertainly. In \cite{B2} the authors present decentralized autonomous exploration method which takes into account frontiers detected by other agent and inter robot exchange of frontiers. This method uses distance and information gain matrices for frontier utility computation.

 \section{METHODOLOGY}
\label{sc: methodology}
 While many research works have been focused on collaborative strategies for SLAM, or single-robot active-SLAM, only a few works have dealt with AC-SLAM. However, these approaches present common limitations: a) they have high computational costs associated with the number of frontiers processed. b) They fail to encourage the spread of robots into the environment by not taking into account the common frontier points from the agents. c) The uncertainty is quantified by a scalar mapping of the entire pose graph covariance matrix which may become very large, especially in landmark-based SLAM methods increasing the computational cost. d) In case of SLAM failure or high uncertainty, there is no re-localization method. Furthermore, they do not explicitly implement strategies for efficient management of frontiers to speed up map discovery and robot localization. In this work, we propose a visual AC-SLAM approach that deals with overcoming these limitations using a fleet of ground and aerial robots and exploiting visual features from different viewpoints of both heterogeneous robots. Our proposed method outlines a strategy aimed at reducing the number of frontiers for reward computation and distributing robots within the environment, thereby facilitating exploration and mapping. Furthermore, we also present a re-localization method quantifying the SLAM pose graph edge covariance matrix hereby allowing the robots to visit already visited goal positions for re-localization and uncertainty reduction for a better SLAM estimate.  We leverage a combination of reward metrics, common map points, and merged map information gain to refine the goal selection.

In the context of single robot A-SLAM the work of \cite{julio1} uses the ORBSLAM2\cite{orbslam2} as SLAM Back-end and proposes a modern D-Optimality criterion for utility function reward computation as the maximum number of spanning trees in the weighted graph Laplacian. 

\begin{figure*}[t]
    \centering
    \includegraphics[height=6cm ,width=15cm]{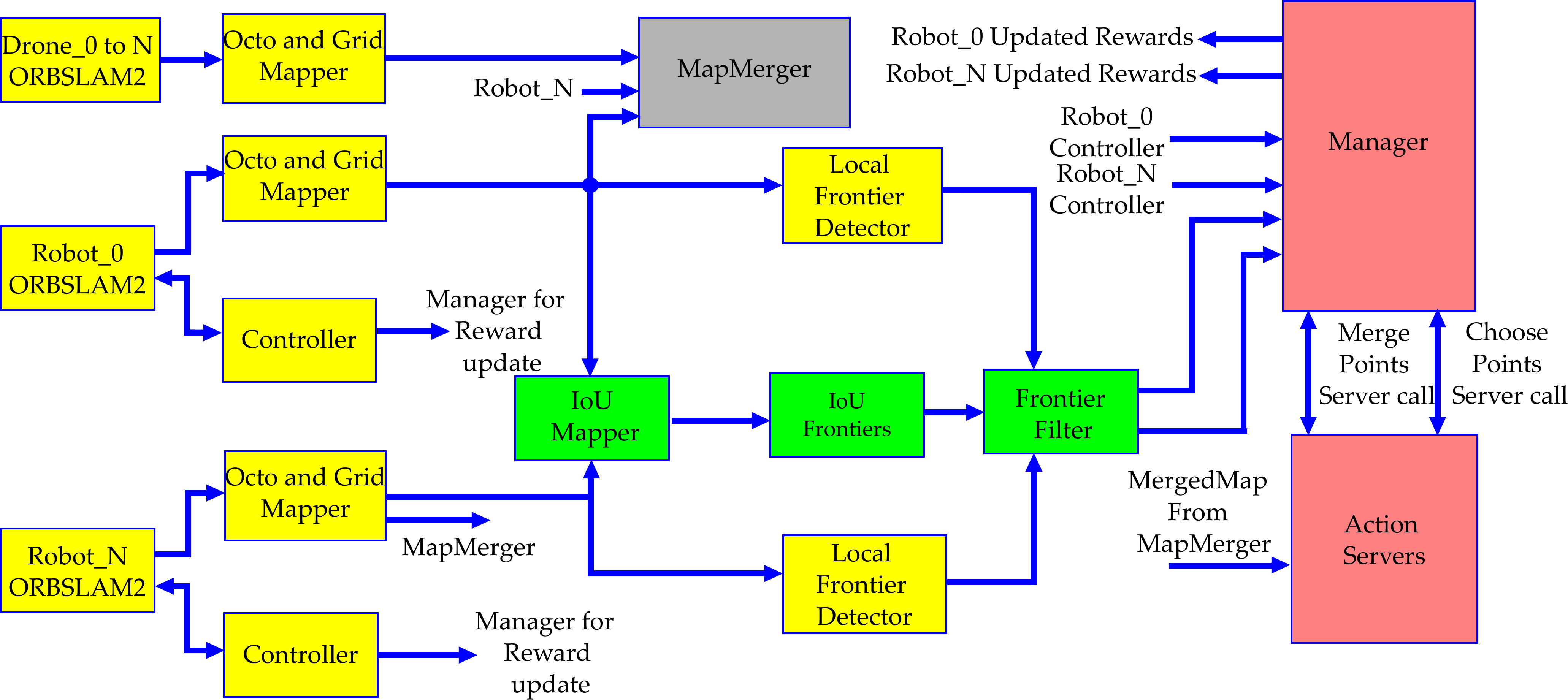}
    \caption{The architecture of resultant system with local (yellow), IoU (green), server (red) and MapMerger (gray) nodes.}
    \label{fig: architecture}
\end{figure*}

We expand the method of \cite{julio1} to a multi-robot AC-SLAM and incorporate the utility function from our previous work in \cite{ahmed1} which considers frontier path entropy to exploit map uncertainty for utility computation. Utilizing the Intersection over Union (IoU) overlapping map information, the number of frontiers is significantly reduced for efficient frontier sharing and exploration. Utilizing edge D-Optimality as an uncertainty metric, we propose a re-localization scheme to encourage loop closure and reduce uncertainty. In Figure \ref{fig: architecture} we show the resultant architecture of the proposed system implemented in ROS Noetic using actionlib\footnote[1]{\url{http://wiki.ros.org/actionlib}} library for UAV and ground robots (Robot\_0... Robot\_N). 
Each robot/UAV runs its own instance of ORBSLAM2 using onboard RGBD sensors. The resultant ORB  features are converted to a 3D voxel map (Octomap) \cite{octomap} for efficient representation using Octo and Grid Mapper module\footnote[2]{\url{http://wiki.ros.org/octomap_server}}. For each of the ground robot and drone, we limit the maximum and minimum height of the Octomap to 0.5 and 1 meters ($m$) respectively, thus limiting the ground robots to ignore ORB features above 0.5$m$ high and for drones to ignore them of less than 1$m$. We then project this Octomap to a 2D Occupancy Grid (O.G) map and apply frontier detection methods using OpenCV and RRT-exploration from \cite{julio1} to detect local frontiers in each robot's local map. The MapMerger module merges the offline drone O.G maps with those of ground robots with world reference frame set to that of Robot\_0.
The IoU mapper computes the overlapped O.G map between two robots and detects frontiers and penalizes each robot's local frontiers thus reducing them. Section \ref{subsec: frontiers_management} elaborates this process. Once the frontiers have been filtered they are passed to the frontier management and reward computation server as discussed in Section \ref{Frontier and reward management Server} which performs further reduction and updates the rewards before finally assigning goal frontiers to the local controller nodes of robots. In the following sections, we describe the functionality of these modules.  

\subsection{IoU frontiers Management}
\label{subsec: frontiers_management}
The local frontiers from each agent are reduced by comparing them with IoU map frontiers using distance based metrics before sending them to the manager node for reward computation. IoU mapper module computes the map using Algorithm \ref{Compute IoU} which takes the O.G maps from Robot\_0 and Robot\_1 as $M1$, $M2$ and returns the IoU map. From Algorithm \ref{Compute IoU}, after the identification of overlapping map coordinates we populate the map occupancy values according to the known free and obstacle information as 0 and 100 respectively. This map is passed to the IoU frontiers module which uses OpenCV frontier detector incorporating Canny edge detection \cite{canny} to detect local frontiers. The frontier filter module takes into account the local frontiers and reduces them by comparing them with those of the IoU map. Algorithm \ref{Frontier Filter} shows an example for Robot\_0 but can be generalized to Robot\_N. It takes as input Robot\_0 local points as 2D map points \text{M1\_pts}, IoU map points \text{IoU\_pts} and a inter point distance threshold \texttt{DIST\_THRSH} parameter adjusted at runtime  for penalizing local points which are too close to those of IoU map.

\begin{algorithm}
\caption{Compute IoU}\label{Compute IoU}
\begin{algorithmic}[1]
\Function{ComputeIoU}{$M1$, $M2$}
    \State $w,h \gets \text {width and height of IoU region}$
    
    \ForAll{$h$ \text{ and } $w$}            
            \State $wx,wy \gets \text{grid to world coord.} $
            \State $idx1, idx2 \gets \text{world coord. to grid index}$
            \State $idx \gets \text{starting index for result.map}$
            \If{$[idx1] \text{ and } [idx2] \neq -1$}
                \If{$[idx1] \land [idx2] = 0$}
                    \State $result.map[idx] \gets 0 $
                \ElsIf{$[idx1] \land [idx2] = 100$}
                    \State $result.map[idx] \gets 100$
                    \ElsIf{$[idx1] \lor [idx2] = 100$}
                    \State $result.map[idx] \gets 100$
                \EndIf
            \EndIf  
    \EndFor

    \State \Return $result.map$
\EndFunction
\end{algorithmic}
\end{algorithm}

\begin{algorithm}
\caption{Frontier Filter}\label{Frontier Filter}
\begin{algorithmic}[1]
\Function{front\_fil}{\text{M1\_pts,}\text{IoU\_pts,} \texttt{DIST\_THRSH}}
    \State $all\_pts \gets \text{M1\_pts} + \text{IoU\_pts,}$
    \State $filtered\_pts \gets \emptyset$
    \ForAll{$p \text{ in } all\_pts$}
        \State $too\_close \gets \text{False}$
        \ForAll{$fp \text{ in } filtered\_pts$}
            \If{$\text{dist}(p, fp) < \texttt{DIST\_THRESH}$}
                \State $too\_close \gets \text{True}$
                \State \textbf{break}
            \EndIf
        \EndFor
        \If{$ \text{not } too\_close$}
            \State \text{add} $p$ \text{ to } $filtered\_pts$
        \EndIf
    \EndFor   
    \State \Return $filtered$
\EndFunction
\end{algorithmic}
\end{algorithm}

\subsection{Frontier and reward management Server}
\label{Frontier and reward management Server}
This section incorporates the \textit{central server} in our previous work of \cite{farhan1} and will be briefly described here. The \textit{central server} comprises of \textit{manager node} and \textit{action servers} that receive the list of frontier points from the frontier filter module. It quantifies them with the merged map information gain and returns the next target goal point to the robot, it also keeps track of already assigned goal points. As shown in Figure \ref{fig: architecture} \textit{manager node} receives the filtered points from each agent and acts as a communication gateway between the server and robot. The \textit{merge points} action server then creates a unique list of frontier points (which is less than the maximum points \texttt{MAX\_PTS}) to be used by all the agents alongside their information gain value depending on the radius \texttt{RAD} and percentage of unknown cells inside it as \texttt{PRC\_UNK}. The \textit{choose goals} action server determines the best goal position for each agent by considering the distance to previously assigned goal points as \texttt{DIST\_THRES} and depending on the reward matrix computation.       
\subsection{Re-localization method}
\label{subsec: Re-localization method}

Pose graph SLAM uses a bipartite graph where each node represents the robot or landmark pose and each edge represents a pose to pose or pose to landmark measurement. The objective is to find the optimal state vector $x^*$ which minimizes the measurement error $e_i(x)$ weighted by the covariance matrix $\Omega_i \in \mathbb{R}^{l \times l}$ where $l$ is the dimension of the state vector $x$ as shown in Equation \ref{slam:eq1}.
 \begin{equation} 
\label{slam:eq1}
	\mathit{x}^{*} = \arg \min_{x} \sum_{i} \mathbf{e}_i^{T}(\mathit{x})\Omega_i\mathbf{e}_i(\mathit{x})
\end{equation}
 The scalar mapping of $\Omega_i$ is used to quantify the uncertainty by reasoning over its Eigenvalues and determinant as described in \cite{carrillo_comparison_2012}. As the robot explores the environment its uncertainly increases and without loop closure or localization help from neighboring robots, it loses its SLAM estimate. We quantify the uncertainty (D-Optimality \cite{carrillo_comparison_2012}) as edge D-Optimality (D-Opti) as defined in Equation \ref{slam:eq2} where $\lambda_1 ,....., \lambda_{k}$ are the Eigenvalues of $\Omega_i$ and $n=6$ is the dimension of each edge covariance matrix. 
\begin{equation} 
\label{slam:eq2}
	\text{D-Opti} = \exp{(\log{(\det(\prod_{k=1,...l}\lambda_{k}))})}/n 
\end{equation}

 To maintain the D-Opti and reduce the SLAM uncertainty, we propose a method to re-localize the robot by guiding it to a previously visited goal point to favor loop closure and improve localization. As shown in Algorithm \ref{Saved Goal Selection Based on Entropy} and implemented in \textit{controller node}. Algorithm \ref{Saved Goal Selection Based on Entropy} takes input the saved goal list as SG\_list, ORBSLAM2 status, D-Opti, \texttt{D\_MAX} is maximum allowed D-Opti value set at runtime and the current robot position R\_pos. Whenever the agent loses localization (ORBSLAM2 status=lost) or D-Opti is greater than \texttt{D\_MAX}, the robot is guided to revisit already visited goal points by weighing them with path entropy as computed in our previous work in \cite{ahmed1}. The robot is assigned the goal point with the largest path entropy and re-localization counter $reloc$ is incremented. 

\begin{algorithm}
\caption{Saved Goal Selection Based on Entropy}\label{Saved Goal Selection Based on Entropy}
\begin{algorithmic}[1]
\Require SG\_list, ORB\_Stat, D-Opti, \texttt{D\_MAX}, R\_pos

\If {(\text{ORB\_Stat is \textbf{lost}}) $\lor$ ($\text{D-Opti} > \texttt{D\_MAX}$)}
    \For {all $item$ in SG\_list}
        \State $ent \gets \text{entropy}(item_x, item_y, \text{R\_pos})$
        \State $\text{egoal\_list}  \gets (1 - ent)$
    \EndFor
    \State $win_{x,y} \gets \text{Max. value in egoal\_list}$, 
    \State \text{send} $win_{x,y}$ \text{to robot}
    \EndIf
    \State $reloc \gets reloc + 1$, \text{SG\_list} $\gets win_{x,y}$ 
\end{algorithmic}
\end{algorithm}

\section{SIMULATION RESULTS}
\label{RESULTS}
\begin{figure}
    \centering
      \subfloat[H.E\label{fig: plugin_map:1a}]{%
           \includegraphics[height=3cm ,width=4.1cm]{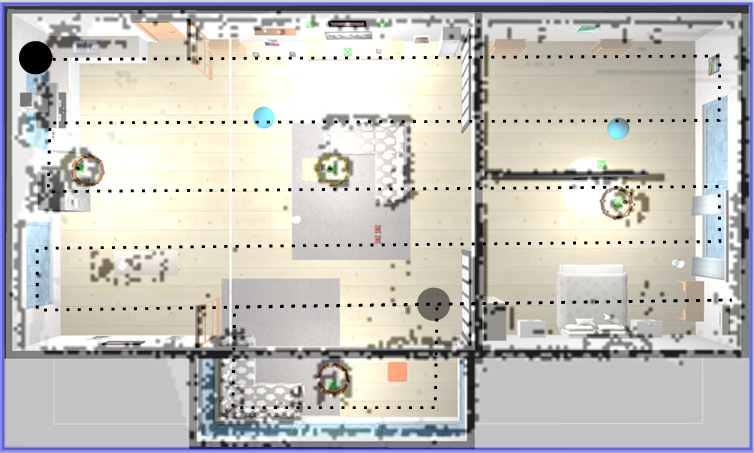}}
        \hfill
      \subfloat[W.E \label{fig: plugin_map:1b}]{%
            \includegraphics[height=3cm ,width=4.4cm]{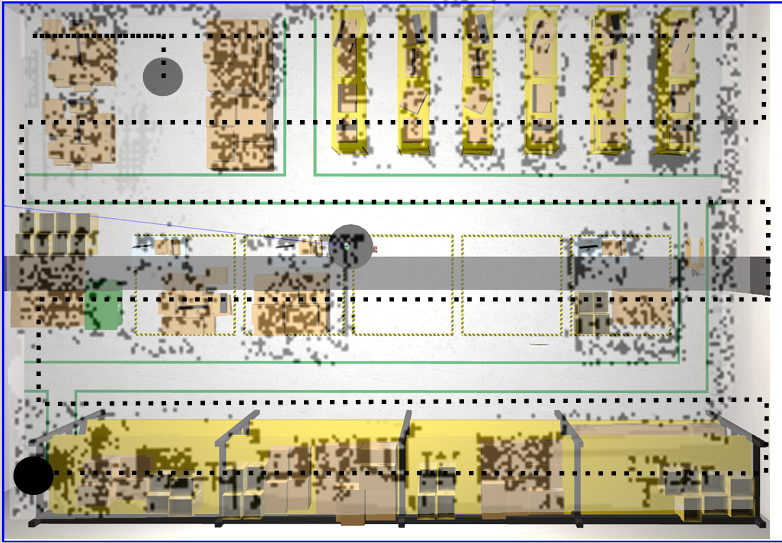}}   

      \subfloat[Resulting merged O.G and Octo-map of H.E. Initial, final robot positions (circles), filtered frontiers (squares) for robot0 (red) and robot1 (green).\label{fig: plugin_map:1c}]{%
           \includegraphics[height=4.7cm ,width=8.6cm]{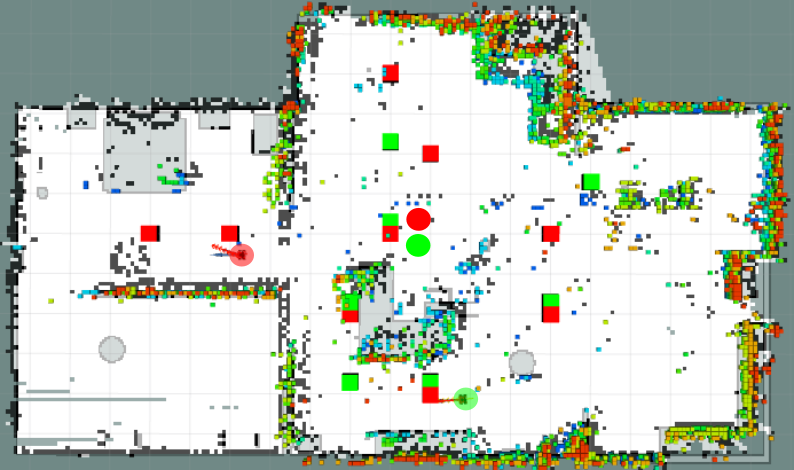}}      
            
      \caption{\ref{fig: plugin_map:1a} and \ref{fig: plugin_map:1b} environments used and the overlapped resulting O.G map from UAV indicating path (black dots), start(black circle), end (gray circle) positions. \ref{fig: plugin_map:1c} showing resulting merged O.G and Octo-map of the H.E.}
      \label{fig: plugin_map}
    \end{figure}

The simulations\footnote[1]{\url{https://www.youtube.com/watch?v=6j3VBdnVcO8}.}. were carried out on ROS Noetic, Ubunto 20.04LTS on Intel Core i7\textsuperscript{\textregistered}, with a system RAM of 32GB and NVIDIA RTX 1000 GPU. As described earlier, we modified the approach of \cite{julio1} to multi-robot and implemented the proposed approach as mentioned in Section \ref{sc: methodology} using ORB SLAM2 \cite{orbslam2} backend, RosBot\footnote[2]{\url{https://husarion.com/}.} equipped with RGBD sensor, LiDAR, and using Timed Elastic Band (TEB) and A$^*$ as local and global planners respectively from the ROS navigation stack \footnote[3]{\url{http://wiki.ros.org/navigation}.}.
We used open-source modified House Environment (H.E) and Warehouse Environment (W.E) from Gazebo simulator\footnote[4]{\url{https://github.com/aws-robotics}.} measuring 157$m^2$ and 260$m^2$ respectively. Figure \ref{fig: plugin_map} shows the Gazebo images masked on top of the resulting O.G maps from UAV by flying it at 5$m$ height in a lawn mover pattern and the final map of H.E. From the O.G maps in Figure \ref{fig: plugin_map:1a} and \ref{fig: plugin_map:1b} we can observe a difference in obstacle area as both environments have different nature of obstacles contributing to more ORB features for W.E. Figure \ref{fig: plugin_map:1c} shows the resulting merged map from Robot\_0, Robot\_1 and UAV,  and indicating the initial, final robot positions alongside their filtered frontiers. The ground truth O.G maps for local and global planners were generated using the Gazebo 2D Map plugin\footnote[5]{\url {https://github.com/marinaKollmitz/gazebo}.} which uses wavefront exploration. 

\begin{figure}
    \centering
      \subfloat[H.E \label{fig:map_all:1a}]{%
           \includegraphics[height=3.0cm ,width=4.31cm]{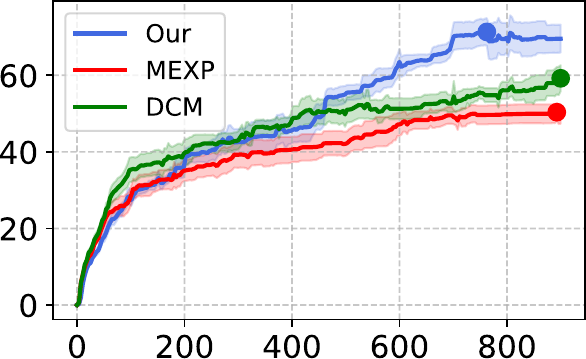}}
        \hfill
      \subfloat[W.E \label{fig:map_all:1b}]{%
            \includegraphics[height=3.0cm ,width=4.31cm]{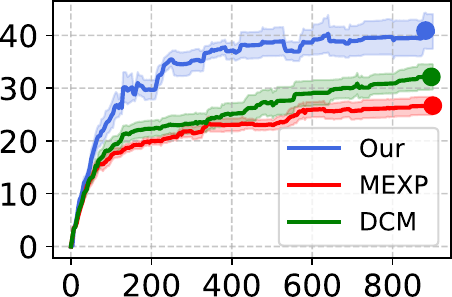}}  
            
      \caption{\% area of map in H.G and W.E environments.}
      \label{fig: map_all}
    \end{figure}

\begin{figure}
    \centering
      \subfloat[H.E\label{fig:map_all2:1a}]{%
           \includegraphics[height=2.8cm ,width=4.31cm]{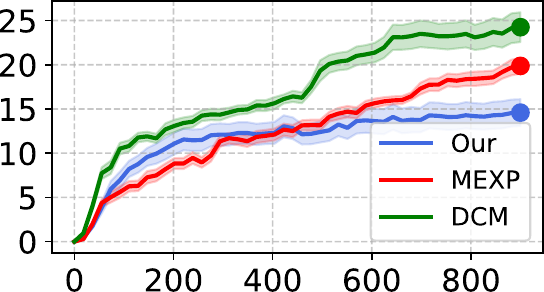}}
        \hfill
      \subfloat[W.E \label{fig:map_all2:1b}]{%
            \includegraphics[height=2.8cm ,width=4.31cm]{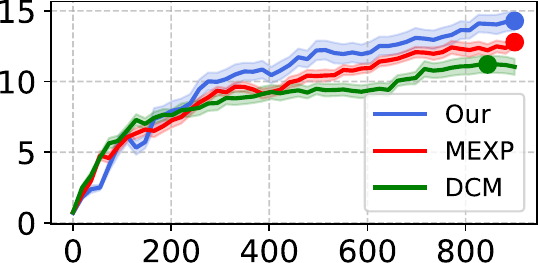}}  
            
      \caption{Area of IoU map in H.E and W.E environments.}
      \label{fig: map_all2}
    \end{figure}

We compared our proposed approach against 1) the approach of \cite{B2} namely DCM which uses inter and intra-agent frontier distance based information gain metrics for frontier utility computation using a decentralized frontier management approach for exploration and mapping. This method does not consider SLAM uncertainty quantification nor does it use any frontier reduction method alongside IoU map consideration. 2) And that of \cite{julio1} by converting it into a multi-robot system namely MEXP without any frontier reduction, reward update, or IoU map consideration as presented in Section \ref{sc: methodology}. For environment exploration we compare metrics of percentage of area covered, IoU map area, frontiers reduction, edge D-Optimality, and number of re-localization efforts. Regarding map quality, we compared metrics measuring Structural Similarity Index Measurement (SSIM) $\in \{0,1\}$, Mean Square Error (MSE) $\in \mathbb{R}$, Normalized Cross Covariance (NCC) $\in \{0,1\}$ and Cosine Similarity (CS)$\in \{0,1\}$ with reference to ground truth maps. These metrics debate over the pixel values, intensity, and structure of the resulting map then compared to the ground truth map.  We conducted 10 simulations of 15 minutes each using 2 robots and 1 UAV for both H.G. and W.E. using Our, MEXP, and DCM methods rendering a total simulation time of 15 hours. The runtime parameters of \texttt{PER\_UNK}, \texttt{RAD}, \texttt{MIN\_PTS}, \texttt{MAX\_PTS}, \texttt{DIST\_THRESH} and \texttt{D\_MAX} were initialized to 60\%, 1$m$, 0, 5, 1$m$ and 1.5 respectively.
Figure \ref{fig: map_all} shows the average percentage of maps discovered along with the Standard Deviation (SD) using our, MEXP, and DCM approaches with reference to time in seconds (time[s]). From Figure \ref{fig:map_all:1a} it can be observed that our approach covers 20\% and 12\% more area in H.E when compared to MEXP and DCM approaches. While in Figure \ref{fig:map_all:1b} Our approach covers 16\% and 11\% more area than MEXP and DCM due to efficient frontiers filtering and utility function. We can also observe that DCM performs better (with more coverage) than MEXP because of the reduced computational time resulting in no uncertainty quantification and its dependence on frontiers information only.
Figure \ref{fig: map_all2} shows the average IoU map area and SD explored in both environments with reference to time[s]. From both Figure \ref{fig:map_all2:1a} and \ref{fig:map_all2:1b} we can conclude that our approach effectively bounds the IoU map area to 15$m^2$ for both environments effectively spreading the robots for exploration.

 \begin{figure}
    \centering
      \subfloat[H.E\label{fig:map_all3:1a}]{%
           \includegraphics[height=2.8cm ,width=4.31cm]{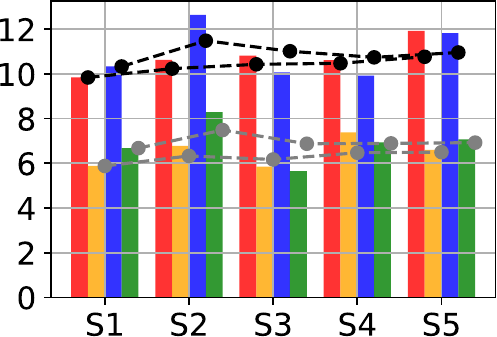}}
        \hfill
      \subfloat[W.E \label{fig:map_all3:1b}]{%
            \includegraphics[height=2.8cm ,width=4.31cm]{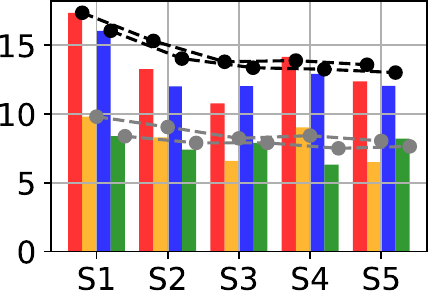}}          
      \caption{Robot\_0 all points (red), Robot\_1 all points (blue), Robot\_0 points after IoU filtering (orange), Robot\_1 points after IoU filtering (green), running average all points (black dots), running average IoU filtered points (gray dots). }
      \label{fig: map_all3}
    \end{figure}

\begin{figure}
    \centering
    \includegraphics[height=4.2cm ,width=8.5cm]{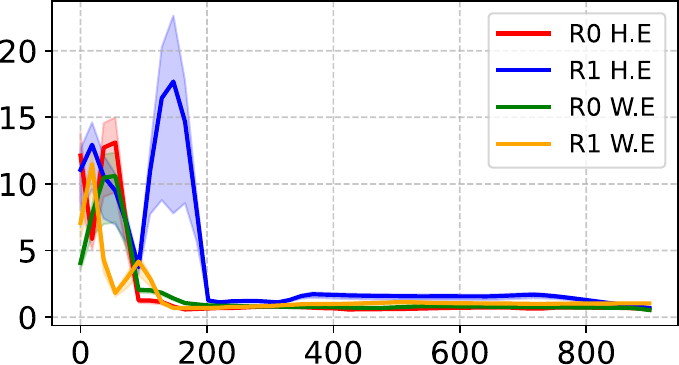}
    \caption{Evolution of edge D-Optimality in environments for Robot\_0 (R0), Robot\_1 (R1).}
    \label{fig: e-dopti}
\end{figure}

\begin{table}
		\centering
  	\caption{Number Of Re-localization efforts.}
    
		\begin{tabular}{|c|c|c|c|c|c|c|c|c|c|c|}
        \hline
        \multicolumn{11}{|c|}{House Environment} \\
        \hline        
        \textbf{Robot} &  \textbf{S1} & \textbf{S2} & \textbf{S3} &\textbf{S4}&\textbf{S5} &\textbf{S6}&\textbf{S7}&\textbf{S8}&\textbf{S9}&\textbf{S10}\\
        \hline  
        \multirow{1}{*}{R0}
               & 4 & 0 & 3 & 2 & 0 & 0 & 2 & 0 & 0 & 0 \\	
        \hline
        \multirow{1}{*}{R1}
              & 0 & 3 & 1 & 3 & 0 & 0 & 0 & 3 & 6 & 0\\	
        \hline
        \multirow{1}{*}{Total}
             & 4 & 3 & 4 & 5 & 0 & 0& 2 & 3& 6 & 0 \\	
        \hline        
        \multicolumn{11}{|c|}{Warehouse Environment} \\
        \hline
        \multirow{1}{*}{R0}
          & 4 & 0 & 1 & 2 & 1 & 2 & 1 & 0 & 0 & 2 \\	
        \hline
        \multirow{1}{*}{R1}
            & 1 & 2 & 0 & 0 & 1 & 3 & 2 & 0 & 0 & 0 \\	
         \hline
        \multirow{1}{*}{Total}
             & 5 & 2 & 1 & 2 & 2 & 5 & 3 & 0 & 0 & 2 \\	
        \hline 
		\end{tabular}		
        \label{tb: d-opti}
	\end{table}

Figure \ref{fig: map_all3} displays the comparison of frontier points reduction in H.E and W.E using our approach along with the running average before and after reduction in black and gray lines for five simulations (S1, S2, S3, S4, S5). From Figure \ref{fig:map_all3:1a} we can deduce that for both R0 and R1 the number of points has reduced to 60\% for H.E while Figure \ref{fig:map_all3:1b} shows a 66\% reduction in W.E. This significant reduction in frontier points consequently reduces the computational cost required by the reward processing on the server side with the adoption of frontier management strategies in Section \ref{subsec: frontiers_management} to limit the number of global frontiers.

When considering the D-Opti and SLAM covariance matrix uncertainty quantification. Figure \ref{fig: e-dopti} shows the evolution of edge D-Opti with reference to time[S]. Initially, all the robots have high uncertainty which eventually decreases as robots explore the environment and due to the efforts to maintain the edge D-Opti to $\leq$\texttt{D\_MAX} using methods described in Section \ref{subsec: Re-localization method}. We can infer that in both environments our method can effectively bound the uncertainty by re-localizing the robots to previously visited goal positions. Table \ref{tb: d-opti} shows the number of re-localization efforts performed by each agent in H.E and H.E. We conclude that in H.E, R1 has performed 16 re-localization efforts to reduce the high uncertainty as evident from Figure \ref{fig: e-dopti} to finally converge to $\leq$\texttt{D\_MAX}. In W.E the robots perform fewer re-localization efforts due to the greater environment ORB features from the UAV O.G map, resulting in more obstacle area as compared to H.E.

\begin{figure}
    \centering
      \subfloat[Our\label{fig:goals:1a}]{%
           \includegraphics[height=2.5cm ,width=2.8cm]{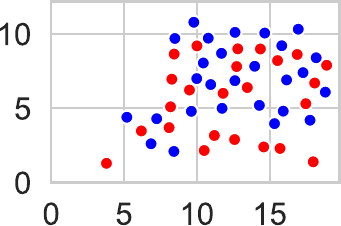}}
        \hfill
      \subfloat[MEXP\label{fig:goals:1b}]{%
            \includegraphics[height=2.75cm ,width=2.8cm]{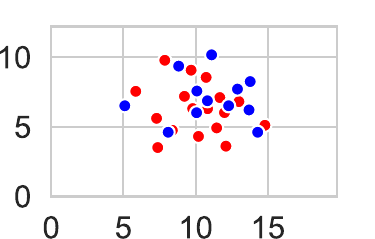}} 
        \hfill
      \subfloat[DCM\label{fig:goals:1c}]{%
            \includegraphics[height=2.7cm ,width=2.8cm]{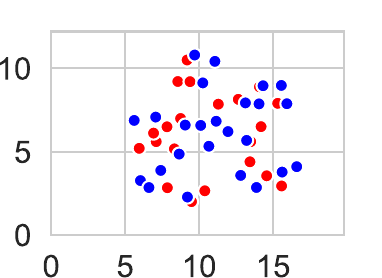}}             
           
      \caption{Spread of final goal positions for H.E R0 (red), R1 (blue).}
      \label{fig: goals}
    \end{figure}

Figure \ref{fig: goals} shows the spread of assigned goal positions with reference to the width and height of H.E map. From Figure \ref{fig:goals:1a} when compared to \ref{fig:goals:1b} and \ref{fig:goals:1c}, we can conclude that due to our efficient frontier management and re-localization policy, we can assign goal positions which support exploration while maintaining efficient SLAM estimate. MEXP in Figure \ref{fig:goals:1b} has no frontier management method hence fewer goal positions due to the high computational cost associated with many frontiers. While in Figure \ref{fig:goals:1c}, DCM manages to have more goal points but they are very close or overlapped to each other, this is because it does not consider inter goal distance while computing frontier information gain.  

Regarding the visual analysis of the maps and debating on map quality metrics in Table \ref{tb: accuracy_comparison}. in almost all cases, using our method rendered reduced MSE and increased SSIM, NCC, and CS as compared to MEXP and DCM methods. We further conclude that the DCM method when compared to MEXP explores the environment more as shown in Figure \ref{fig: map_all}, but has higher MSE and lower SSIM, NCC, and NC.  

\begin{table}
		\centering
  	\caption{MAP QUALITY METRICES.}
   		\begin{tabular}{|c|c|c|c|c|c|c|}
        \hline
        \textbf{Env} & \textbf{Method} & \textbf{MSE} & \textbf{SSIM} &\textbf{NCC}&\textbf{CS} \\
        \hline        
        \multirow{1}{*}{H.E}
              & Our & \textbf{7544.238}  &  \textbf{0.386} & \textbf{0.426} & \textbf{0.455} \\	
        \hline
        \multirow{1}{*}{H.E}
              & MEXP & 7673.455  &  0.308 & 0.228 & 0.214 \\  \hline
        \multirow{1}{*}{H.E}
          & DCM & 8695.255  &  0.198 & 0.125 & 0.155 \\		
        \hline
        \multirow{1}{*}{W.E}
             & Our &  \textbf{8753.571}  & \textbf{0.352} & \textbf{0.415} & \textbf{0.347} \\	       	
        \hline
        \multirow{1}{*}{W.E}
              & MEXP & 10160.746  &  0.265 & 0.344 &  0.262 \\	
         \hline
        \multirow{1}{*}{W.E}
              & DCM & 11059.335  & 0.139 & 0.117 &  0.061 \\	
         \hline
		\end{tabular}		
        \label{tb: accuracy_comparison}
	\end{table}
   
\section{CONCLUSIONS}
\label{sc: conclusions}
We proposed a method for the coordination of multiple robots in a collaborative exploration domain performing visual AC-SLAM. We proposed a strategy to efficiently reduce the number of frontiers for the agents to compute their reward functions to reduce the computational cost and to spread the robots into the environment. We also proposed a re-localization method to promote loop closure. We presented extensive simulation analysis on publicly available environments and compared our approach to similar methods and achieved to explore an average of 32\% and 27\% more area. Possible future works can explore exchanging the local pose graphs for collaborative localization and mapping. 

\addtolength{\textheight}{-8.5cm}   

\section*{ACKNOWLEDGMENT}
This work was conducted within the framework of the NExT Senior Talent Chair DeepCoSLAM, funded by the French Government through the program "Investments for the Future" administered by the National Agency for Research (ANR-16-IDEX-0007). We also extend our gratitude to the Région Pays de la Loire and Nantes Métropole for their invaluable support in facilitating this research endeavour. 


\bibliographystyle{IEEEtran}
\bibliography{main}
\end{document}